\documentclass[letterpaper, 10pt, conference]{ieeeconf}

\makeatletter
\let\NAT@parse\undefined
\makeatother

\usepackage{amssymb}
\usepackage[T1]{fontenc}
\usepackage[utf8]{inputenc}
\usepackage[final, stretch=10, shrink=50]{microtype}
\usepackage{booktabs}
\usepackage[rgb]{xcolor}
\usepackage{graphicx}
\usepackage[american]{babel}
\usepackage{relsize}
\usepackage[per=slash]{siunitx}
\usepackage{inconsolata}
\usepackage{xspace}
\usepackage{csquotes}
\usepackage[style=ieee, backend=biber, url=false]{biblatex}
\usepackage[hidelinks]{hyperref}
\usepackage{subfigure}
\usepackage{tikz}

\pdftrailerid{}
\AtBeginDocument{
	\AtEveryBibitem{\clearfield{doi}}
	\AtEveryBibitem{\clearfield{series}}
	\AtEveryBibitem{\clearfield{issn}}
	\AtEveryBibitem{\clearfield{isbn}}
	\AtEveryBibitem{\clearfield{month}}
	\AtEveryBibitem{\clearfield{day}}
	\AtEveryBibitem{\clearlist{language}}
	\AtEveryBibitem{\clearfield{note}}
	\AtEveryBibitem{\ifentrytype{book}{}{\clearname{editor}}}
}
\defbibenvironment{bibliography}
{\list
	{\printtext[labelnumberwidth]{%
			\printfield{prefixnumber}%
			\printfield{labelnumber}}}
	{\setlength{\labelwidth}{\labelnumberwidth}%
		\setlength{\leftmargin}{0.64cm}%
		\setlength{\labelsep}{0.2cm}%
		\setlength{\itemsep}{\bibitemsep}%
		\setlength{\parsep}{\bibparsep}}%
	}
{\endlist}
{\item}
\IEEEoverridecommandlockouts
\DeclareSIUnit\px{px}
\DeclareSIUnit\fps{fps}
\addto\extrasamerican{}
\addto\extrasamerican{}
\addto\extrasamerican{}
\newcommand{\etal}{\textit{et~al.}}
\newcommand{\cpp}{C\nolinebreak[4]\hspace{-.05em}\raisebox{.4ex}{\relsize{-3}{\textbf{++}}}}
\newcommand{\yolo}{\textsc{yolo}}
\newcommand{\rgb}{\textsc{rgb}}
\newcommand{\rgbd}{\textsc{rgb\babelhyphen{nobreak}d}}
\newcommand{\cad}[1]{\textsc{cad\babelhyphen{nobreak}#1}}

\newcommand{\crf}{\textsc{crf}}
\newcommand{\vr}{\rule[-.35\baselineskip]{0pt}{\baselineskip}}
\newcommand{\semec}{\textsc{sec}}
\newcommand{\ie}{i.\,e.,\ }
\newcommand{\eg}{e.\,g.,\ }
\newcommand{\armarx}{\mbox{ArmarX}\xspace}

\title{\LARGE \bf%
Learning Object-Action Relations from Bimanual Human Demonstration Using Graph Networks%
}

\author{Christian R.\,G. Dreher, Mirko Wächter, and Tamim Asfour%
\thanks{The research leading to these results has received funding from the European Union’s Horizon 2020 Research and Innovation programme under grant agreement No 643950 (SecondHands) and the Carl Zeiss Foundation.}
\thanks{The authors are with the High Performance Humanoid \mbox{Technologies} (H$^2$T) lab, Institute for Anthropomatics and Robotics (IAR), \mbox{Karlsruhe} Institute of Technology (KIT), Karlsruhe, Germany,
{\tt \{\href{mailto:Christian\%20Dreher<c.dreher@kit.edu>?subject=Learning\%20Object-Action\%20Relations\%20from\%20Bimanual\%20Human\%20Demonstration\%20Using\%20Graph\%20Networks}{c.dreher}, waechter, asfour\}@kit.edu}.}%
}

\begin{document}

\maketitle
\thispagestyle{empty}
\pagestyle{empty}

\begin{abstract}

Recognizing human actions is a vital task for a humanoid robot, especially in domains like programming by demonstration.
Previous approaches on action recognition primarily focused on the overall prevalent action being executed, but we argue that bimanual human motion cannot always be described sufficiently with a single action label.
We present a system for frame-wise action classification and segmentation in bimanual human demonstrations.
The system extracts symbolic spatial object relations from raw \rgbd\ video data captured from the robot's point of view in order to build graph-based scene representations.
To learn object-action relations, a graph network classifier is trained using these representations together with ground truth action labels to predict the action executed by each hand.

We evaluated the proposed classifier on a new \rgbd\ video dataset showing daily action sequences focusing on bimanual manipulation actions.
It consists of 6 subjects performing 9 tasks with 10 repetitions each, which leads to 540 video recordings with 2 hours and 18 minutes total playtime and per-hand ground truth action labels for each frame.
We show that the classifier is able to reliably identify (action classification macro $F_1$-score of 0.86) the true executed action of each hand within its top 3 predictions on a frame-by-frame basis without prior temporal action segmentation. 

\end{abstract}

\begin{figure}[t!]
    \def\svgwidth{\linewidth}
	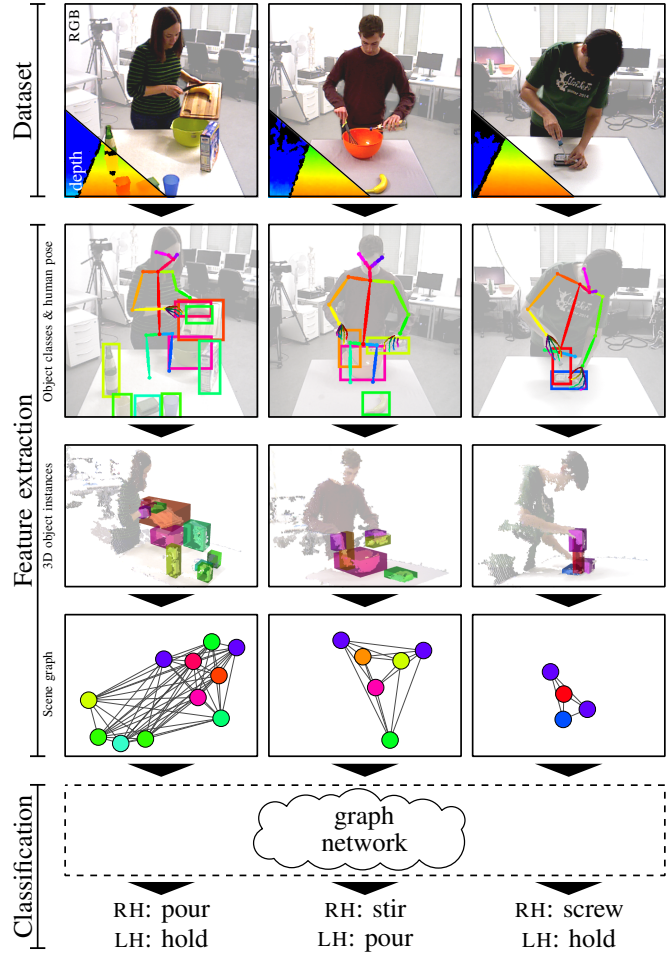
	\caption{Simplified overview and outline of our contributions in context. \emph{Dataset}: 3 exemplary \rgbd{} images from the dataset (images edited for better clarity); \emph{Feature extraction}: The interim results and final result of the 3-stage processing pipeline, a scene graph; \emph{Classification}: A graph network classifier making predictions about the performed action for the right hand (\textsc{rh}) and the left hand (\textsc{lh}) based on the scene graph.}
	\label{fig:intro}
	\vspace*{-0.4cm}
\end{figure}

\section{Introduction}

For domains like programming by demonstration \cite{billard2008robot}, it is vital for a robot to be able to recognize the actions of a human.
The obtained information can be used to learn action sequences from a human teacher in order to replicate tasks, or anticipate what a human wants to do to timely assist them.
Most previous approaches (\eg \cite{koppula2016anticipating, ziaeetabar2018recognition, ji20133d, kjellstrom2011visual, rodriguez2008action, blank2005actions, aksoy2011learning, wachter2015hierarchical, mandery2015analyzing}) on action recognition usually assigned one single action label to each point in time, but we argue that this is not enough in general, considering natural bimanual human motion.
Take, for example, a baking task where one has to fold egg whites into a dough.
This implies two actions, as one is required to \emph{pour} egg whites into a bowl with one hand, while \emph{folding} them in with the other.
This simple example is not easily representable with a single label, but assigning an action label to each hand solves this.
Especially bimanual programming by demonstration approaches~\cite{zollner2004programming} could benefit from this granularity of information in order to individually discriminate the semantic role of each hand.

One important aspect of programming by demonstration is the question of how to form an abstract knowledge representation.
Using symbolic features for that has several benefits, as raw video data streams are of high dimensionality and have no inherent semantic meaning.
Additionally, this provides the desired abstraction layer, both for the representation, as well as for the elicitation of the symbolic features.
An example for this representation is a robot observing a human teacher who pours water from a bottle into a cup.
It is more general to memorize the scene in a symbolic way (\ie while \emph{pouring}, the \emph{bottle} is \emph{above} the \emph{cup}) instead of determining exact coordinates of the corresponding objects.
In this work, we focus on 3D symbolic spatial relations between the human hands and the objects for each given point in time, and represent this scene as a graph.
Object-action relations are learned by training a graph network~\cite{battaglia2018relational} classifier, a machine learning building block, with those scene graphs.
In particular, the classifier learns to estimate action labels for each hand given a history of scene graphs.
In order to evaluate, which spatial relations between a given pair of objects and the human hands are in effect, \rgbd\ video data was used.

To conclude this introduction, the main contributions of our work (presented in \autoref{s:approach}) are as follows:
\begin{itemize}
	\item A novel \rgbd{} video dataset specifically tailored to research bimanual human actions.
	\item A pipeline to construct scene graphs from \rgbd{} videos.
	\item A frame-wise action segmentation and recognition approach, which is invariant to the number or order of object instances, does not require a prior temporal segmentation, and predicts actions for each hand individually by learning object-action relations. 
\end{itemize}
\autoref{fig:intro} shows the contributions in context. The whole \rgbd\ dataset, as well as supplementary and derived data, are publicly available at \href{https://bimanual-actions.humanoids.kit.edu}{\texttt{bimanual-actions.humanoids.kit.edu}}.

\begin{figure*}[t]
    \def\svgwidth{\linewidth}
	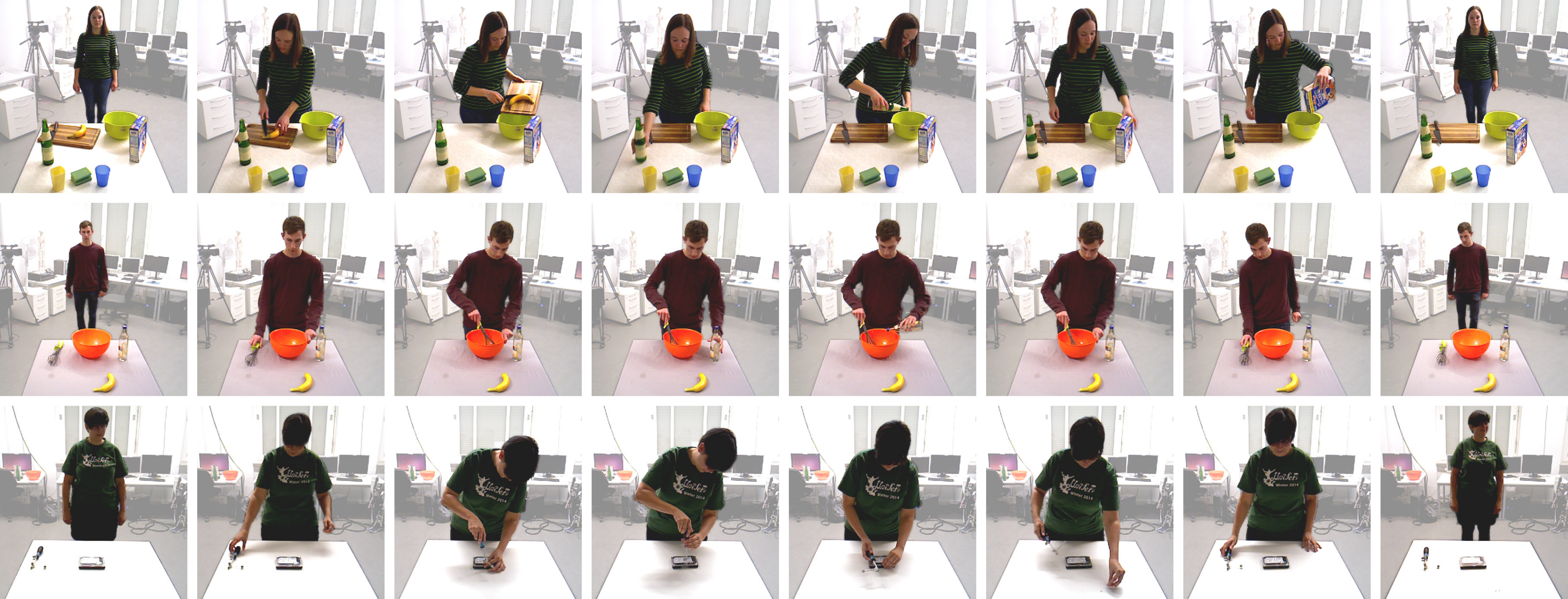
    \vspace*{-5mm}
	\caption{Exemplary recordings from our proposed dataset. First row: Preparing breakfast cereals by cutting and pouring a banana into a bowl, followed by milk and cereals. Second row: Cooking by stirring in a bowl while pouring water from a bottle to it. Third row: Disassembling a hard drive by unscrewing and removing a screw. (Images cropped/edited and depth images omitted for the sake of clarity and brevity.)}
	\label{fig:dataset}
	\vspace*{-5.5mm}
\end{figure*}

\section{Related Work}

In this section, multiple related works are considered regarding their datasets, and their methods towards predicting human actions from various input modalities.

\subsection{Datasets}

There are several video datasets compiled to research human action recognition problems, where the recording modalities range from \rgb{} only \cite{kuehne2014language, damen2018scaling, zhou2018automatic} over \rgbd\ \cite{wu2015watchnpatch, sung2011human, wang2012mining, xia2012view, koppula2013learninghuman, aksoy2015modelfree} to complex multimodal environments \cite{tenorth2009tum}.

Kühne~\etal~\cite{kuehne2014language} presented a large \rgb\ dataset of cooking activities, called the \emph{Breakfast Actions Dataset}, with a total length of over \SI{77}{\hour}.
Damen~\etal~\cite{damen2018scaling} compiled the \textsc{epic kitchens} dataset, where subjects were asked to wear a head-mounted GoPro to record their cooking activities.
With vast amounts of videos being publicly available on the internet, Zhou~\etal~\cite{zhou2018automatic} collected \SI{176}{\hour} of instructional videos for their YouCook2 dataset.
The selected videos were temporally segmented into procedure steps, but mostly contain cuts.
Since we wanted to evaluate 3D spatial relations between objects, we could not make use of any of these datasets.
Apart from that, they were not suitable for our scenario where a robot observes a human teacher, because they either do not provide a viewing angle as seen from the robot, or are not continuous in time.

Wu~\etal~\cite{wu2015watchnpatch} compiled a large \rgbd\ dataset for action recognition.
It features 7 subjects in 458 high quality recordings and has a total length of about \SI{3}{\hour} \SI{50}{\minute}.
Other \rgbd\ datasets from the Cornell University are the \cad{60} with Sung~\etal~\cite{sung2011human}, and the \cad{120} with Koppula~\etal~\cite{koppula2013learninghuman}.
The datasets differ in the granularity of the annotation, and in size, as the \cad{120} is twice as large.
In both datasets, the camera angle relative to the subject varies.
Wang~\etal~\cite{wang2012mining} collected a dataset of activities of daily living, recorded from a fixed \rgbd\ camera in front of a sofa. 
Xia~\etal~\cite{xia2012view} had 10 subjects perform 10 indoor activities in front of a fixed camera.
Aksoy~\etal~\cite{aksoy2015modelfree} introduced the \textsc{maniac} dataset, which features 5 subjects over 140 \rgbd\ videos in total.
Most of these datasets were recorded using a Microsoft Kinect.
All of them, however, did not suit our needs, mostly because of the viewing angles, their small number of recordings, or their focus on activities rather than on fine-grained actions.

There are also other approaches, like the \textsc{tum} Kitchen dataset, collected by Tenorth~\etal~\cite{tenorth2009tum}.
It was recorded in an intelligent kitchen with several \rgb\ cameras mounted on the ceiling and other kinds of sensors, and they considered the left hand and right hand separately for the ground truth.
But again, our focus lies in the sensors available on a humanoid robot in a one-on-one scenario.
With the exception of this dataset, none of the others discussed here, regardless of the modalities, considered bimanual actions, but instead focused on the overall prevalent activity or action.

For more extensive comparisons we refer to Poppe~\cite{poppe2010survey}, Weinland \etal~\cite{weinland2011survey}, or Chaquet \etal~\cite{chaquet2013survey}, where most of these datasets were discussed in great detail.
Additionally, Zhang~\etal~\cite{zhang2016rgbdbased} specifically surveyed \rgbd\ datasets for action recognition.

\subsection{Action Recognition}

Similar to the datasets, also the action recognition approaches can be divided into those who use \rgbd\ data \cite{koppula2016anticipating, ziaeetabar2018recognition}, \rgb\ data only \cite{kjellstrom2011visual, blank2005actions, rodriguez2008action, ji20133d}, and others \cite{tenorth2009tum, aksoy2011learning, wachter2015hierarchical, mandery2015analyzing}.

In many cases, conditional random fields (\crf{}s) were employed, a probabilistic graphical machine learning approach \cite{koppula2016anticipating, tenorth2009tum, kjellstrom2011visual}.
Koppula and Saxena~\cite{koppula2016anticipating} used a \crf\ to classify action segments, and therefore heavily relied on a prior accurate temporal segmentation.
Kjellström~\etal~\cite{kjellstrom2011visual} used a \crf\ to learn object-action relations.
Their method simultaneously classified and segmented actions, but only considered one hand.
Tenorth~\etal~\cite{tenorth2009tum} considered both hands, but evaluated only on the left hand.
All of these approaches model their problem in a chained graph structure, which is required so that the inference on \crf{}s is feasible.

Some early approaches interpret an action as a spatio-temporal volume of image frames over time, extracting the shape of the action by subtracting the background \cite{blank2005actions, rodriguez2008action}.
In a more modern interpretation of this approach, Ji~\etal~\cite{ji20133d} used a 3D convolutional neural network to not only convolve spatially, but also temporally.

Aksoy~\etal~\cite{aksoy2011learning} coined the term of \emph{Semantic Event Chains} (\semec{}s), a concept which encodes transitions between object relations in a matrix. 
The work on \semec{}s was further continued by Ziaeetabar~\etal~\cite{ziaeetabar2018recognition}, where \semec{}s were enriched with a large array of static and dynamic spatial relations.
In order to evaluate the spatial relations, 3D bounding boxes estimating the objects were used, calculated from \rgbd{} images.

Wächter and Asfour~\cite{wachter2015hierarchical}, as well as Mandery~\etal~\cite{mandery2015analyzing}, used the change of contact relations as strong indicator for the presence of temporal segmentation boundaries.

Action recognition is a common problem, and therefore there are a multitude of other approaches available.
For a broader overview on older methods not mentioned here, we again refer to the works of Poppe~\cite{poppe2010survey} or Weinland \etal{}~\cite{weinland2011survey}.
More recent approaches focusing on \rgb{} input data are discussed by Herath~\etal{}~\cite{herath2017going}.

Except for that of Tenorth~\etal~\cite{tenorth2009tum}, none of these works consider natural bimanual actions in the sense that each hand may perform an individual action, like stirring in a bowl while pouring water in it.
Additionally, the used machine learning approaches often limit the application to a fixed set of object instances and to a specific order.
We are only bound to a fixed set of object \emph{classes}, multiple instances can easily be represented in the scene graph.
Other than that, we do not require any prior temporal segmentation.

\begin{figure*}[t!]
  \centering
  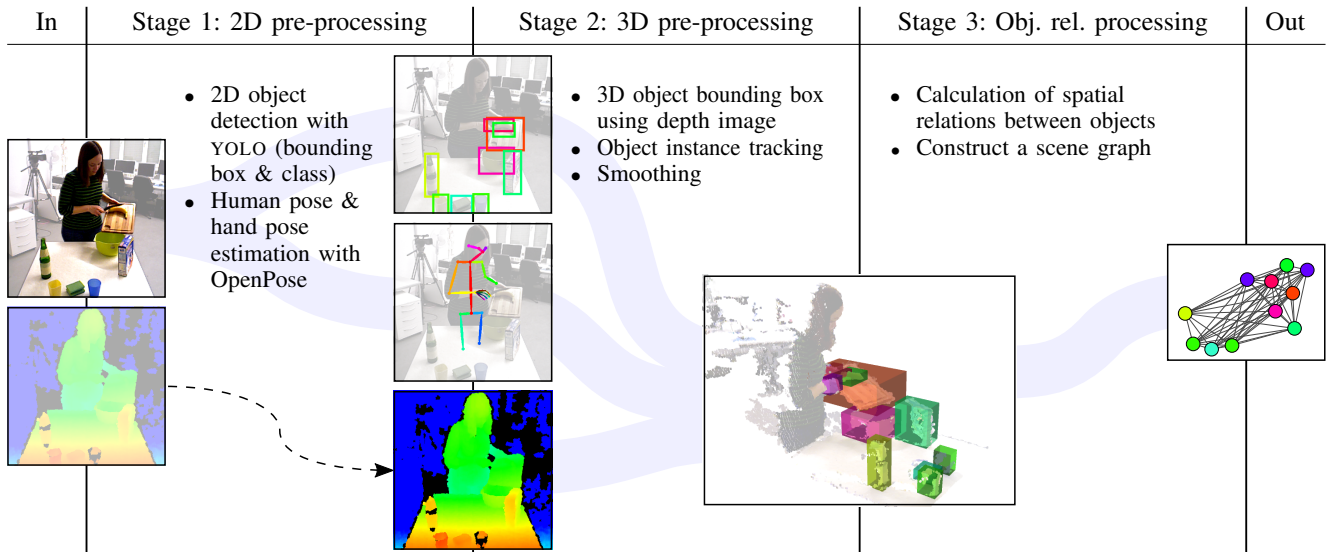
  \vspace*{-3mm}
  \caption{Schematic of the 3-stage processing pipeline.
  Input: An \rgbd{} image. Output: A scene graph.
  Stage 1: 2D pre-processing, computing the 2D object bounding boxes and the human pose from the \rgb{} image, forwarding the depth image to stage 2.
  Stage 2: 3D pre-processing, computing 3D object bounding boxes, tracking object instances, and smoothing the noise resulting from the depth image.
  Stage 3: Object relation processing, evaluating which spatial relations between each pair of objects are in effect, constructing a scene graph.
  For each stage, the inputs are depicted on the left border and the outputs on the right.}
  \label{fig:pipeline}
  \vspace*{-5mm}
\end{figure*}

\section{Approach}\label{s:approach}

In this section, we will present our proposed approach for the segmentation and recognition of bimanual actions by learning the relation between objects and actions. 
The 3 contributions, also depicted in \autoref{fig:intro}, are described in detail in the following subsections.
First, we describe the \rgbd\ dataset collected to train the developed classifier in \autoref{sec:dataset}. 
In \autoref{sec:featex}, we present a feature extraction pipeline, which takes such \rgbd\ data as input and constructs a scene graph.
The nodes in such a scene graph encode the object classes (including hands), and an edge encodes the relations between two objects.
The objects are detected using an object detection framework, while the hands are detected using a human pose estimation framework, both taking \rgb\ images as input.
Finally, in \autoref{sec:ac-recog}, we introduce the main contribution of this paper.
To learn object-action relations, \ie the relation between a scene graph and the executed action, we employ a graph network classifier, a type of machine learning building block designed to operate on variable-sized graphs.

\subsection{Dataset}\label{sec:dataset}

For this work, a rich \rgbd{} video dataset of bimanual action sequences was compiled, 3 of which are shown exemplarily in \autoref{fig:dataset} in a few key frames.
We recorded 6 subjects (3 female, 3 male; 5 right-handed, 1 left-handed) performing 9 different tasks (5 in a kitchen context, 4 in a workshop context) from a robot's point of view.
The considered tasks were 1. and 2. \emph{cooking} in two variants (pour from bottle vs. pour from bowl), 3. \emph{pouring water}, 4. \emph{wiping the table}, and 5. \emph{preparing breakfast cereals} for the kitchen tasks, as well as 6. and 7. \emph{disassembling a hard drive} in two variants (hard drive on the table vs. in the hand), 8. \emph{hammering nails}, and 9. \emph{sawing wood} for the workshop tasks.
Each task was repeated 10 times.
This totals to 540 recordings of fully labeled bimanual actions with a total runtime of approx. \SI{2}{\hour} \SI{18}{\minute}.
More precisely, one annotator manually labeled the whole dataset frame-wise once for each hand with one of 14 possible action classes in $A = \{$\,\emph{idle}, \emph{approach}, \emph{retreat}, \emph{lift}, \emph{place}, \emph{hold}, \emph{stir}, \emph{pour}, \emph{cut}, \emph{drink}, \emph{wipe}, \emph{hammer}, \emph{saw}, \emph{screw}\,$\}$.
Wächter and Asfour~\cite{wachter2015hierarchical} used a similar detail of labeling in which the hand \emph{approaches} an object and, after using it, \emph{retreats}.
In most cases, the object also has to be \emph{lifted} and \emph{placed} for usage (\eg \emph{pouring}).

Apart from the objects the subject interacted with, up to 3 additional known and contextually fitting objects were placed on the table.
The 12 considered object classes are $O = \{$\,\emph{cup}, \emph{bowl}, \emph{whisk}, \emph{bottle}, \emph{banana}, \emph{cutting board}, \emph{knife}, \emph{sponge}, \emph{hammer}, \emph{saw}, \emph{wood}, \emph{screwdriver}\,$\}$.
For the classes \emph{cup}, \emph{bottle}, \emph{bowl}, and \emph{sponge}, several differently looking objects were used, as can be seen in \autoref{fig:dataset}.
Additionally, 5413 frames (about 10 random frames per recording) were manually labeled with object class bounding boxes by the same annotator. This data was later used to train an object detection framework.

The hardware used to record the dataset was a PrimeSense Carmine 1.09 \rgbd{} camera, which captures images at \SI{30}{\fps} with a resolution of \SI{640x480}{\px}.
It was attached to a tripod at a height of \SI{1.7}{\meter} to simulate a standing robot.
The camera was tilted forth, so that the entire working space and the teacher's head were still pictured.
Due to technical limitations at the time of recording the first subject, 83 of the 540 recordings had to be captured at only \SI{15}{\fps}.

\vspace*{-1mm}

\subsection{Feature Extraction}\label{sec:featex}

The feature extraction is a vital link in our work to convey \rgbd{} images to scene graphs, on which our classifier operates on.
We chose to work with symbolic features as opposed to following an end-to-end approach, since this greatly reduces the problem dimensionality.
This means that less data is required, but also that these symbolic features need to be extracted first.
To do so, a 3-stage pipeline was deployed.
It was implemented in \cpp{} and integrated into the \armarx{} framework~\cite{vahrenkamp2015robot}.
\autoref{fig:pipeline} shows a schematic of the whole processing pipeline.
To give a brief overview: The input of the pipeline are consecutive \rgbd{} images, however only one \rgbd{} image is fed into the pipeline per pass.
While the \rgb{} image is used in the first stage, the depth image is not needed until the second stage.
The output of the pipeline is a scene graph, where all detected object instances are represented as nodes, and all present relations between each pair of objects are encoded into the edges.
Each pass of the pipeline takes about \SI{250}{\milli\second} with an Nvidia \textsc{titan x}.
The outputs of all stages were recorded for every frame and are available for download as well.
The following paragraphs will describe each stage in more detail.

\vspace*{-0.1mm}

The first stage in the pipeline is the 2D pre-processing, where the objects are detected with \yolo{}~\cite{redmon2018yolov3} (trained on the objects in our dataset) and the hands of the human teacher with OpenPose~\cite{cao2018openpose} by feeding the \rgb{} image.
The hand key points provided by OpenPose are used to calculate the 2D bounding box for each hand.
Hence, this stage outputs a list of 2D bounding boxes of the objects detected by \yolo{} and the hands detected by OpenPose.
Note that the hands are treated as any object in the following.

\vspace*{-0.1mm}

The second stage performs the 3D pre-processing, where the data of the first stage is used in conjunction with a point cloud derived from the depth image to acquire the 3D bounding boxes of the objects.
This is achieved by clustering only that part of the point cloud, which is outlined by the 2D bounding boxes, and the assumption, that the biggest cluster in terms of point count belongs to the detected object.
Minimum and maximum extents of that cluster yield the 3D bounding box.
Since the depth images suffer from high-frequency noise, which directly transfers to the 3D bounding boxes, this stage also performs a smoothing by applying a Gaussian filter over the parameters of the past observed 3D bounding boxes of each object.
The Gaussian filter was parameterized so that $3\sigma = \SI{250}{\milli\second}$.
Furthermore, to be able to apply the smoothing and to later calculate dynamic spatial relations between objects, it is important to identify concrete object instances over several frames.
This stage therefore also includes an object tracking algorithm.
The output of this stage are 3D object bounding boxes enriched by globally unique object instance identifiers.

The third stage is the object relation processing. The 3D bounding boxes of the previous stage are used to determine, which of the spatial relations are present for a given pair of objects.
We considered the 15 spatial relations from Ziaeetabar~\etal~\cite{ziaeetabar2018recognition}, namely $R = \{\,$\emph{contact}, \emph{above}, \emph{below}, \emph{left}, \emph{right}, \emph{front}, \emph{behind}, \emph{inside}, \emph{surround}, \emph{moving together}, \emph{halting together}, \emph{fixed moving together}, \emph{getting close}, \emph{moving apart}, \emph{stable}\,$\}$.
Contrary to their formulation, however, no exception conditions were used.
The output of this stage, and therefore the pipeline, is a scene graph, where nodes represent object instances, and edges encode spatial relations between them.

\begin{figure*}[t!]
  \centering
  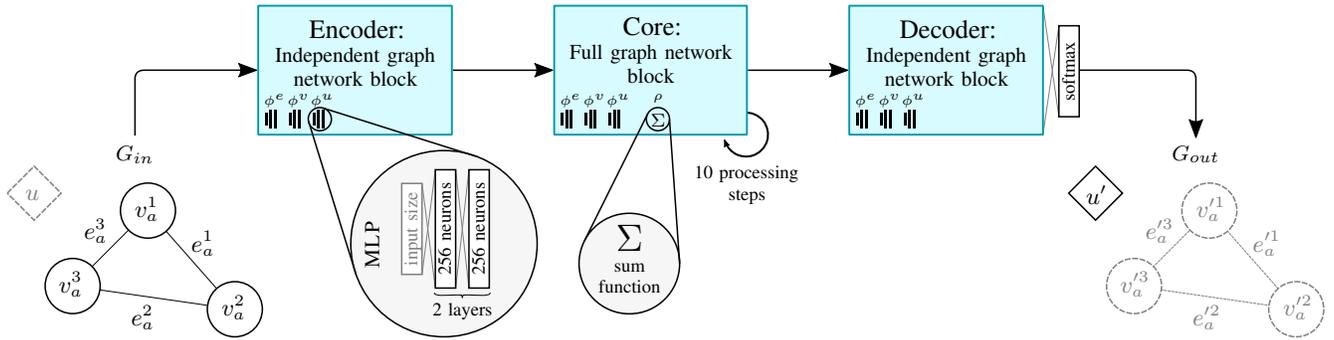
  \vspace*{-2mm}
  \caption{Architecture of our action classifier, an encode-process-decode graph network with 10 processing steps.
  The input is a scene graph $G_{in}$ (here exemplarily with 3 nodes) with edge attributes $e_a$ (relations), and node attributes $v_a$ (objects).
  The global attribute $u'$ of the output graph $G_{out}$ encodes the predicted probability distribution of actions.
  Grayed out attributes are not used by our classifier.
  Both the encoder and the decoder, as well as the core, use 3 instances of a multilayer perceptron (\textsc{mlp}) parameterized as depicted for each of their update functions $\phi^e$, $\phi^v$, and $\phi^u$.
  The core uses the sum function for each aggregation function $\rho \in \{\rho^{e\rightarrow{}v}, \rho^{v\rightarrow{}u},  \rho^{e\rightarrow{}u} \}$.
  An additional layer is applied right after decoding to scale the latent size to the actual size of $u'$ on the one hand, and apply a softmax to get a probability distribution on the other.}
  \label{fig:arch}
  \vspace*{-5mm}
\end{figure*}

\subsection{Classification}\label{sec:ac-recog}

\vspace{-0.3mm}

To learn object-action relations from \rgbd{} videos, we employed a graph network classifier \cite{battaglia2018relational} together with the scene graphs returned from our feature extraction pipeline.
Graph networks are machine learning building blocks operating on attributes which can be arranged as a graph.
Battaglia~\etal~\cite{battaglia2018relational} define a graph $G$ as a 3-tuple $G = (u, V, E)$, where $u$ is the global attribute of the graph, $V$ the set of nodes in the graph, and $E$ the set of edges.
Each $v_a \in V$ is a node attribute and each $e \in E$ is a 3-tuple $e = (e_a, s, r)$.
In this, $e_a$ is the edge attribute, and $s$ and $r$ are the indices of the sender and receiver node in $V$.
A graph network takes such a graph as input, processes it by updating its attributes, and returns it afterwards.
The processing takes place in 3 steps, in which following functions are applied: (1) An edge update function $\phi^e$; (2) an edge aggregation function $\rho^{e\rightarrow{}v}$ and a node update function $\phi^v$; (3) one aggregation function for the nodes $\rho^{v\rightarrow{}u}$ and one for the global attribute $\rho^{e\rightarrow{}u}$, as well as a global update function $\phi^u$.
This describes a full graph network block, but different types of blocks are possible depending on which update or aggregation functions are used.
For example, in an independent graph network block, no aggregation functions are used.
Graph networks can also be composed from several graph network blocks, they do not necessarily have to be atomic graph network blocks.
Again, different configurations are possible here as well.
For more details about graph networks, we refer to the original publication~\cite{battaglia2018relational}.

We used the \emph{encode-process-decode} configuration for our model, depicted in \autoref{fig:arch}, for which a reference implementation is available in the Graph Nets library~\cite{battaglia2018relational}.
This model consists of two independent graph network blocks for the encoder and decoder respectively, and a full graph network block for the core.
For all 3 blocks, multilayer perceptrons (\textsc{mlp}s) were employed as edge update functions $\phi^e$, node update functions $\phi^v$, and global update functions $\phi^u$.
For the aggregation functions $\rho^{e\rightarrow{}v}$, $\rho^{v\rightarrow{}u}$, and $\rho^{e\rightarrow{}u}$, the sum function was used.
All \textsc{mlp}s in each graph network block were parameterized with 2 layers and 256 neurons per layer.
The core in the encode-process-decode model performed 10 processing steps.
These parameters were empirically determined after evaluating multiple test series, each sampling a different configuration.
The input of our classifier is a scene graph $G_{in}$, where the edge attributes $e_a$ encode the relations and the node attributes $v_a$ the object classes.
The output is a probability distribution of all actions, which is encoded in the updated global attribute $u'$.
The global attribute $u$ and the updated edge and node attributes $e'_a$ and $v'_a$ are not used.

In our case, all data is symbolic, so one-hot encodings were used for the actions, objects, and relations.
The global attribute $u$ encodes the performed action of one hand.
This leads to the one-hot encoding $u \in \{0,1\}^{|A| = 14}$ for the 14 considered actions.
The node attributes $v_a \in V$ encode 12 object classes known to \yolo{} and one object class per hand. This leads to 14 object classes in total and $v_a \in \{0,1\}^{|O|+2=14}$.
All relations are encoded as edge attributes $e_a \in \{0,1\}^{|R|+1=16}$, 15 for the spatial relations, plus one to encode a temporal relation.

\begin{figure}[t]
  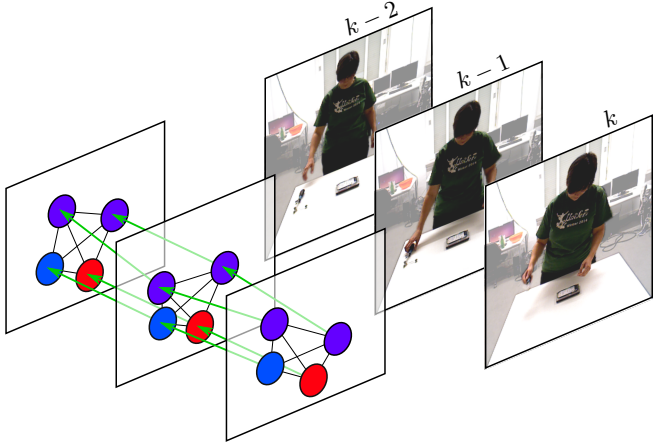
  \vspace*{-6.5mm}
  \caption{Example of how a temporal concatenation of scene graphs (left) is constructed (corresponding video frames of each scene graph to the right).
  For the sake of simplicity, only 3 considered frames ($k-2$, $k-1$, and $k$) are shown.
  The temporal edges are depicted in green and trace one object instance over a series of frames.}
  \label{fig:tmp-scene-graph}
  \vspace*{-6mm}
\end{figure}

Due to noisy depth images and occasional misclassifications from \yolo{}, certain scene graphs might be ambiguous or not representative for the currently performed action in the frame.
To mitigate this effect, we fed a \emph{temporal concatenation} of 10 consecutive scene graphs instead of only one scene graph for the current frame (the current frame plus the 9 previous ones; roughly \SI{333}{\milli\second} at \SI{30}{\fps}).
By temporal concatenation of scene graphs we understand an algorithm, which takes a list of scene graphs as input, and outputs one single scene graph, where all nodes and edges from the input scene graphs are included.
The resulting scene graph's global attribute is adopted from the current scene graph while training, but most importantly, \emph{temporal edges} are supplemented by the algorithm.
These are edges, which connect the nodes of one specific object instance over a series of frames.
\autoref{fig:tmp-scene-graph} shows an illustration of this process.
In other words, a temporal concatenation preserves the number of nodes and edges encoding spatial relations.
All nodes connected by spatial relations always belong to one frame, while a path along temporal edges tracks one object instance over multiple frames.
Therefore, edges for spatial relations and temporal edges are mutually exclusive.
This approach is comparable to how Koppula~\etal~\cite{koppula2013learninghuman, koppula2016anticipating} encode temporal relations, however they use it to connect nodes over temporal segments rather than over frames.

Conceptually, the classifier is trained on solely the right hand.
To account for the left hand, we trained the classifier on \emph{mirrored scene graphs} as well.
A mirrored scene graph is a scene graph, where the objects \emph{right hand} and \emph{left hand}, as well as the relations \emph{right of} and \emph{left of} are swapped.
Additionally, while training, the ground truth in the global attribute of the target graph is changed to the action performed by the left hand.
The benefits of this approach are the reduced training and setup effort, as well as twice as much data through mirroring.
At runtime, the scene graph is fed into the network to obtain the action for the right hand.
Afterwards, the scene graph will be mirrored and fed into the network again, to get the action for the left.

\section{Evaluation}

\vspace*{-0.5mm}

For the following evaluations, we trained the classifier on our new dataset.
For each involved training process, the dataset was split into a training set and a testing set.
Testing sets always contained all recordings from one subject, while training sets contained all remaining ones.
Additionally, before training, 1 out of the 10 repetitions for each task in the training set were put aside as validation set.
We chose a batch size of at most 512 samples, but because of class imbalances, 2 out of 3 samples for the actions \emph{idle} and \emph{hold} were discarded per batch.
The Adam optimizer was used with a learning rate of 0.001, and the loss was defined as the cross entropy of the softmax of the global attribute (right hand action) and the ground truth.
We stopped the training after the graph network started overfitting and used that state for the evaluation on the test set.
The prediction of an action took about 75\,ms on an Intel i7 \textsc{cpu}, but can further be improved by utilizing a \textsc{gpu}.

\definecolor{ac_idle}{RGB}{204,204,204}
\definecolor{ac_approach}{RGB}{255,123,116}
\definecolor{ac_retreat}{RGB}{255,12,0}
\definecolor{ac_lift}{RGB}{153,255,149}
\definecolor{ac_place}{RGB}{0,255,9}
\definecolor{ac_stir}{RGB}{0,88,255}
\definecolor{ac_hold}{RGB}{246,235,135}
\definecolor{ac_pour}{RGB}{151,0,255}
\definecolor{ac_drink}{RGB}{255,134,0}

\newcommand*{\actionsegment}[1]{%
    \setbox0=\hbox{\strut}%
    \begin{tikzpicture}
    \tikzstyle{every node}=[draw]
    \path[shape=coordinate]
    (0,0) coordinate(b1) (0.2,0) coordinate(b2)
    (0.2,0.2) coordinate(b3) (0,0.2) coordinate(b4);
    
    \fill [#1] (0,0) rectangle (0.2,0.2);
    \draw[line width=0.3mm] (b4) -- (b1);
    \draw[line width=0.3mm] (b2) -- (b3);
    \end{tikzpicture}%
}

\begin{figure*}[t!]
    \def\svgwidth{\linewidth}
    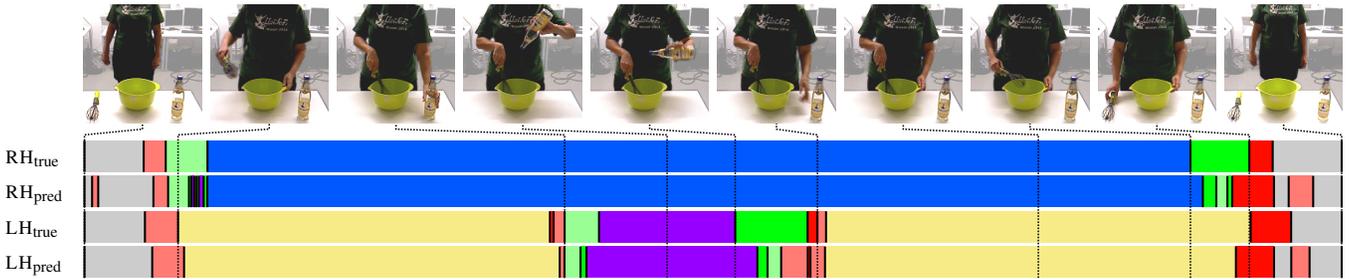
    \vspace*{-6mm}
    \caption[Qualitative evaluation]{Qualitative evaluation by visualizing the top prediction of the classifier for the right hand ({\sc rh}$_\text{pred}$) and left hand ({\sc lh}$_\text{pred}$) in each frame over one whole example recording next to the corresponding ground truth ({\sc rh}$_\text{true}$ and {\sc lh}$_\text{true}$).
        Consecutive predictions of the same class were pooled into an action segment of one color.
        Each color depicts a certain action:
        \actionsegment{ac_idle}~idle,
        \actionsegment{ac_approach}~approach,
        \actionsegment{ac_retreat}~retreat,
        \actionsegment{ac_lift}~lift,
        \actionsegment{ac_place}~place,
        \actionsegment{ac_stir}~stir,
        \actionsegment{ac_hold}~hold, and
        \actionsegment{ac_pour}~pour.
    }
    \label{fig:qual-eva}
\end{figure*}

\begin{figure*}[t!]
    \vspace{-0.4cm}
    \centering
    \subfigure{\includegraphics[trim=0 0 0 4mm, clip, width=.45\textwidth]{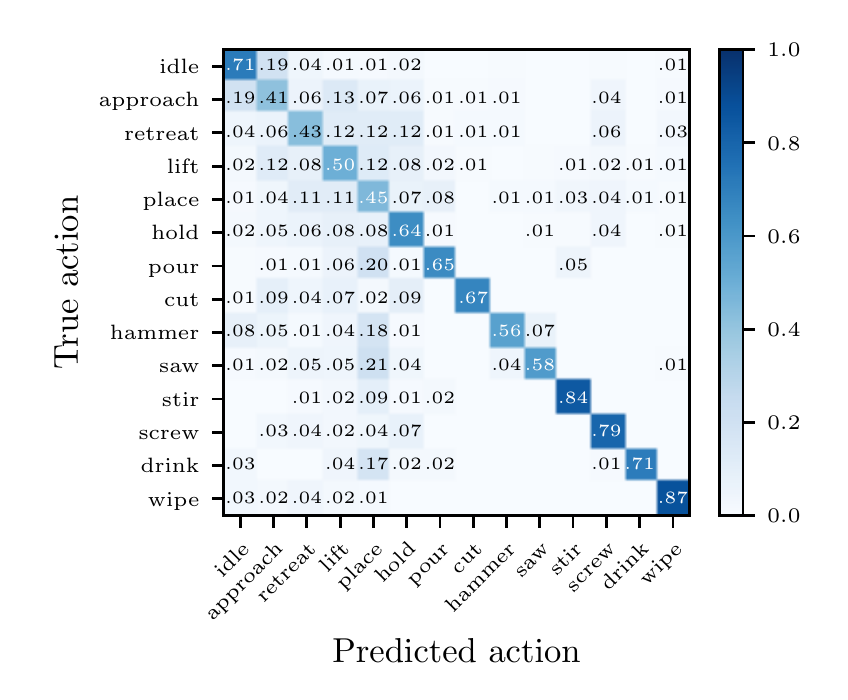}}
    \subfigure{\includegraphics[trim=0 0 0 4mm, clip, width=.45\textwidth]{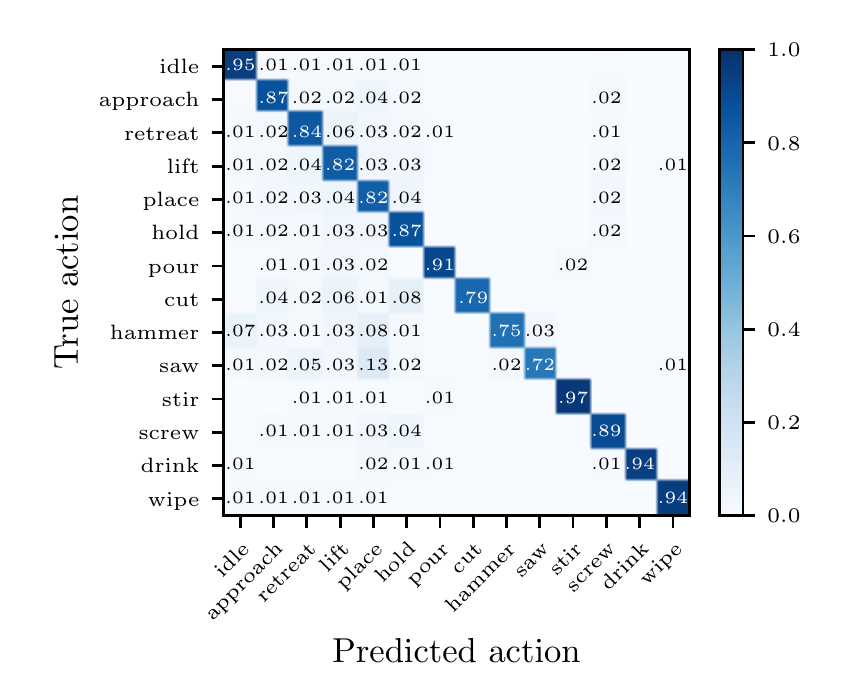}}
    \vspace*{-5.6mm}
    \caption{Accumulative classification correctness over all folds depicted as normalized confusion matrix for the top prediction (left), and where a classification result was considered correct if the ground truth was in the top 3 predictions (right). Empty cells mean no confusion ($= 0.00$).}
    \label{fig:cm}
    \vspace*{-5.5mm}
\end{figure*}

\autoref{fig:qual-eva} shows a qualitative evaluation on a recording from subject 1, where the classifier was trained with recordings of the remaining 5 subjects.
The top predictions for both hands in each frame were determined, and adjacent predictions were pooled into one action segment.
Apart from a few oscillations while lifting the whisk, the classifier was able to form larger contiguous action segments close to the ground truth.

\begin{table}[t]
    \caption{Quantitative evaluation results.}
    {%
        \vspace*{-0.3cm}
        \noindent Precision (Precis.), Recall and $F_1$ score of the action classification once for each action class, as well as Micro, Macro and Weighted (Weigh.) averages (avg.) thereof. Apart from considering the top prediction only, we also evaluated those scores again where a classification result was considered correct if the ground truth was in the top 3 predictions.
        \vspace*{0.2cm}
        \par
    }
    \centering
    \begin{tabular}{ r r r r r r r }
        \toprule[1pt]
        & \multicolumn{3}{c}{Top prediction} & \multicolumn{3}{c}{Top 3 predictions} \\
        \cmidrule{2-4} \cmidrule(l{2pt}){5-7}
        Action class & Precis. & Recall & $F_1$ & Precis. & Recall & $F_1$ \\
        \midrule[0.5pt]
        \emph{idle}     &  0.85 &   0.71 &  0.78 &  0.97 &   0.95 &  0.96 \\
        \emph{approach} &  0.31 &   0.41 &  0.35 &  0.84 &   0.87 &  0.86 \\
        \emph{retreat}  &  0.34 &   0.43 &  0.38 &  0.78 &   0.84 &  0.81 \\
        \emph{lift}     &  0.32 &   0.50 &  0.39 &  0.67 &   0.82 &  0.74 \\
        \emph{place}    &  0.34 &   0.45 &  0.38 &  0.73 &   0.82 &  0.77 \\
        \emph{hold}     &  0.82 &   0.64 &  0.72 &  0.93 &   0.87 &  0.90 \\
        \emph{pour}     &  0.66 &   0.65 &  0.66 &  0.91 &   0.91 &  0.91 \\
        \emph{cut}      &  0.74 &   0.67 &  0.70 &  0.89 &   0.79 &  0.83 \\
        \emph{hammer}   &  0.64 &   0.56 &  0.60 &  0.83 &   0.75 &  0.79 \\
        \emph{saw}      &  0.68 &   0.58 &  0.63 &  0.88 &   0.72 &  0.79 \\
        \emph{stir}     &  0.92 &   0.84 &  0.88 &  0.98 &   0.97 &  0.98 \\
        \emph{screw}    &  0.76 &   0.79 &  0.77 &  0.88 &   0.89 &  0.89 \\
        \emph{drink}    &  0.70 &   0.71 &  0.70 &  0.94 &   0.94 &  0.94 \\
        \emph{wipe}     &  0.78 &   0.87 &  0.82 &  0.92 &   0.94 &  0.93 \\
        \midrule[0.5pt]
        Micro avg.      &  0.64 &   0.64 &  0.64 &  0.89 &   0.89 &  0.89 \\
        Macro avg.      &  0.63 &   0.63 &  \textbf{0.63} &  0.87 &   0.86 &  \textbf{0.86} \\
        Weigh. avg.     &  0.69 &   0.64 &  0.66 &  0.89 &   0.89 &  0.89 \\
        \bottomrule[1pt]
    \end{tabular}
    \label{tab:eva-res}
    \vspace*{-5mm}
\end{table}

For the quantitative evaluation of the classifier, a leave-one-subject-out cross-validation was performed to obtain 6 folds of training and testing sets.
The results of this evaluation are listed in \autoref{tab:eva-res}, once for only the top prediction of the classifier, and once again where a prediction was counted as correct if the ground truth was in the top 3 predictions.
The latter allows to evaluate, how good the classifier was at identifying correct action candidates.
\autoref{fig:cm} depicts the normalized confusion matrices, again for the top prediction only, and for the top 3 predictions.
As can be seen from the action classification macro $F_1$ score of 0.86, the classifier is generally able to reliably identify correct candidates.
The confusion matrix for the top prediction, however, indicates that in certain cases, it lacks important information to discriminate actions.

A major confusion of the classifier was the prediction of \emph{place} while the true action was \emph{saw}, \emph{pour}, \emph{hammer}, or \emph{drink}.
The cause for the prediction of \emph{drink} is that there is no reliable point of reference the classifier could have made use of to distinguish handling a cup (\emph{lifting}, \emph{holding}, or \emph{placing}) from actually \emph{drinking}.
Currently, we only consider the human hands, but adding the head to the scene graph could greatly improve the classifier's performance, as \emph{contact} relations would be enough to reliably detect it.
The confusions with \emph{saw}, \emph{pour}, or \emph{hammer} can be attributed to wrongly detected 3D bounding boxes.
For our feature extraction pipeline, it was especially hard to correctly determine the 3D bounding box for very thin objects like hammers or saws.
This can be contributed to our method of estimating them, namely through clustering the part of the depth image outlined by the 2D bounding box and using the biggest cluster.
This assumption often fails for thin objects, as the background cluster (mostly the abdomen of the subject) yields more points.
Another problem with similar effects was the fact that the bottles (especially the green one) consistently did not yield useful depth information due to high absorption.
The described effect can even be observed in the confusion matrix for the top 3 predictions.
To mitigate this, it could prove beneficial to replace the bounding box object detection approach with one that yields bounding polygons, which would completely eradicate the need to perform a clustering in the first place.
Additionally, often it would have been helpful to consider the table as object as well to improve the distinction of actions like \emph{lift} and \emph{place}, since the dynamic spatial relations \emph{moving apart} or \emph{getting close} with the table as object would be a strong indicator for either one or the other.

There is also a number of confusions noticeable between the contact-less actions \emph{approach} or \emph{retreat} and the contacting actions \emph{lift}, \emph{place}, or \emph{hold}.
The main cause for this is the coarse method to detect contacts, namely collision checks between 3D axis aligned bounding boxes.
Additionally, phases of approaching or retreating are usually executed very fast (sometimes within tenths of a second).
However, such infor\-mation is not directly gathered by our feature extraction at this point.
Both kinds of confusions could be mitigated by using oriented bounding boxes to estimate the extents of objects.

We also performed the evaluation on the top 3 predictions to show that there is still a lot potential to improve and to motivate future work, as the classifier is generally already able to identify correct candidates for the action.
This can be seen in the confusion matrix in \autoref{fig:cm}~(right) and the action classification macro $F_1$ score of 0.86.
As already pointed out, the top confusions here are the ones involving \emph{place} in conjunction with thin objects,
\eg the prediction of \emph{place} when the true action was \emph{saw} or \emph{hammer}.
The comparison to the confusion matrix in \autoref{fig:cm}~(left) clearly suggests, that the classifier would be able to make better predictions if it was provided with more information of better quality to make the final decision.
The suggested improvements to the feature extraction aim at exactly that and could supplement the scene graph in order to provide the classifier with the needed information to resolve the confusions.

To verify the effectiveness of the architecture and the classifier, an ablation study was performed by removing a set of features, training, and evaluating a new model.
First, it was assessed how the classifier performs when only contact relations are considered and all other symbolic spatial relations are discarded.
In this case, the action classification macro $F_1$ score declined to 0.46 compared to the score of 0.63 from \autoref{tab:eva-res}, where all spatial relations were considered.
This shows that other symbolic spatial relations indeed encode important information the classifier can make use of.
Next, the classifier was assessed without considering any spatial relations, and instead encoding the object bounding box centroids in the graph nodes.
This resulted in an action classification macro $F_1$ score of 0.31, showing that the classifier is not able to derive features of similar quality comparable to the symbolic relations, and that \emph{contact} relations alone are valuable information.
Finally, an evaluation on raw scene graphs without any temporal concatenations was performed, to assess, how beneficial they are.
This resulted in an action classification macro $F_1$ score of 0.60.
Even though the classifier performed better with temporal relations, the improvement was rather minor considering the up to 10 times higher processing effort for training and execution.
The results of this ablation study are listed in \autoref{tab:ablation-eva-res}.

\begin{table}[t!]
    \vspace{2mm}
    \caption{Ablation study evaluation results.}
    {%
        \vspace*{-0.3cm}
        \noindent Action classification macro $F_1$ scores for the evaluations performed in the ablation study, namely considering contact relations only (Contact), considering object centroid coordinates instead of symbolic relations (Centroids), and considering no temporal relations (No temp.). The results are compared to those achieved in our quantitative evaluation (Reference), both for the top prediction only (Top pred.) and where a classification result was considered correct if the ground truth was in the top 3 predictions (Top 3 pred.).
        \vspace*{0.2cm}
        \par
    }
    \centering
    \begin{tabular}{ r r r r r }
        \toprule[1pt]
        & Contact & Centroids & No temp. &     Reference \\
        \midrule[0.5pt]
        Top pred.   &    0.46 &      0.31 &     0.60 & \textbf{0.63} \\
        Top 3 pred. &    0.73 &      0.55 &     0.84 & \textbf{0.86} \\
        \bottomrule[1pt]
    \end{tabular}
    \label{tab:ablation-eva-res}
    \vspace*{-5mm}
\end{table}

A direct comparison to approaches from the literature proved challenging, because, to the best of our knowledge, there are no bimanual action recognition approaches to compare with.
Considering only one hand is not meaningful, because this is not the same as the action labels for the overall prevalent action as seen in most other approaches and datasets.
Additionally, \yolo{} currently limits us to our dataset, as we trained it to specifically recognize the objects occurring in that only.

\section{Conclusion and Future Work}

In this work, we presented an approach to learn object-action relations from bimanual human demonstration, for action segmentation and recognition.
The proposed classifier takes scene graphs as input, and provides bimanual predictions of the performed actions.
Using a graph network allows us to encode a scene without having to consider the amount or order of the objects.
Additionally, we do not require any prior action segmentation at this point.
To obtain the scene graphs from \rgbd{} video frames, we developed a feature extraction pipeline making use of two state-of-the-art vision frameworks to detect objects and the human teacher's pose to obtain 3D symbolic spatial relations between objects and hands.
Other than that, we contribute a novel \rgbd{} dataset of subjects performing bimanual actions in the kitchen and workshop.
The ground truth action labels are provided on a per-hand basis.

In future work, including the table as global point of reference and the head of the teacher could further improve the prediction quality.
The pose information of the human teacher is available and could be used to account for the whole upper body movement. 
The biggest negative impact on the prediction quality, however, can be attributed to wrong relations resulting from misplaced 3D bounding boxes.
At this point, we also only consider 10 frames ($\approx \SI{333}{\milli\second}$) to predict the action in the most recent frame, but long term sequence information could be important data for the classifier, as certain actions follow a logical sequence (\eg \emph{lift} implies \emph{place} later on).

\renewcommand*{\bibfont}{\footnotesize}
\setlength\bibitemsep{-0.06mm}
\printbibliography

\end{document}